\documentclass[nohyperref]{article}

\usepackage{microtype}
\usepackage{graphicx}
\usepackage{subfigure}
\usepackage{booktabs} 
\usepackage{fancyhdr}

\usepackage{hyperref}

\usepackage[accepted]{icml2022}

\usepackage{amsmath}
\usepackage{amssymb}
\usepackage{mathtools}
\usepackage{amsthm}

\usepackage[capitalize,noabbrev]{cleveref}

\theoremstyle{plain}

\theoremstyle{definition}

\theoremstyle{remark}

\usepackage[textsize=tiny]{todonotes}

\icmltitlerunning{Published as a conference paper at ICML 2022}

\begin{document}

\twocolumn[
\icmltitle{Error-driven Input Modulation: Solving the Credit Assignment Problem without a Backward Pass}



\icmlsetsymbol{equal}{*}

\begin{icmlauthorlist}
\icmlauthor{Giorgia Dellaferrera}{aaa,yyy,zzz}
\icmlauthor{Gabriel Kreiman}{aaa,yyy}
\end{icmlauthorlist}

$^1$ Center for Brains, Minds and Machines, Cambridge, MA, United States\\
$^2$ Children’s Hospital, Harvard Medical School, Boston, MA, United States \\
$^3$ Institute of Neuroinformatics, University of Zurich and ETH Zurich, Switzerland\\
Correspondence to: Giorgia Dellaferrera $<$giorgia.dellaferrera@gmail.com$>$, Gabriel Kreiman $<$gabriel.kreiman@childrens.harvard.edu$>$ \\

\icmlaffiliation{aaa}{Center for Brains, Minds and Machines, Cambridge, MA, United States}
\icmlaffiliation{yyy}{Children’s Hospital, Harvard Medical School, Boston, MA, United States}
\icmlaffiliation{zzz}{Institute of Neuroinformatics, University of Zurich and ETH Zurich, Switzerland}

\icmlcorrespondingauthor{Giorgia Dellaferrera}{giorgia.dellaferrera@gmail.com}
\icmlcorrespondingauthor{Gabriel Kreiman}{gabriel.kreiman@tch.harvard.edu}

\icmlkeywords{Biologically inspired AI, training scheme, Machine Learning, ICML}

\vskip 0.3in
]




\begin{abstract}
Supervised learning in artificial neural networks typically relies on  backpropagation, where the weights are updated based on the error-function gradients and sequentially propagated from the output layer to the input layer. Although this approach has proven effective in a wide domain of applications, it lacks biological plausibility in many regards, including the weight symmetry problem, the dependence of learning on non-local signals, the freezing of neural activity during error propagation, and the update locking problem. Alternative training schemes have been introduced, including sign symmetry, feedback alignment, and direct feedback alignment, but they invariably rely on a backward pass that hinders the possibility of solving all the issues simultaneously.  Here, we propose to replace the backward pass with a second forward pass in which the input signal is modulated based on the error of the network. We show that this novel learning rule comprehensively addresses all the above-mentioned issues and can be applied to both fully connected and convolutional models. We test this learning rule on MNIST, CIFAR-10, and CIFAR-100. These results help incorporate biological principles into machine learning.
\end{abstract}

\section{Introduction}
The backpropagation algorithm (BP) has proven to reach impressive results in training Artificial Neural Networks (ANNs) on a broad range of complex cognitive tasks including speech recognition, image classification \cite{Lecun_Deeplearning}, and playing board games \cite{Silver}. However, BP has been criticized for relying on a biologically unrealistic strategy of \textit{synaptic credit assignment}, \textit{i.e.,} estimating how much each parameter has contributed to the output error \cite{Crick89,Whittington,Lillicrap2020}. In particular, a few aspects of BP appear to be at odds with neurobiology. (i) The same weights are used both in the feedforward and in the feedback pathway, raising the weight symmetry or \textit{weight transport problem} \cite{Burbank2012}. (ii) The parameter updates depend on the activity of all downstream nodes, while biological synapses learn based on local signals related to the activity of the neurons they connect with \cite{Whittington}. (iii) The error gradients are stored separately from the activations \cite{liao2016signsymmetry} and do not influence the activities of the nodes produced in the forward pass. Hence, during the backward pass, the network activity is frozen. In the brain, instead, the neural activity is not frozen during plasticity changes and the signals travelling through feedback connections influence the neural activities produced by forward propagation, leading to their enhancement or suppression \cite{Lillicrap2020}. (iv) Input signals cannot be processed in an online fashion, but each sample needs to wait for both the forward and backward computations to be completed for the previous sample. This is referred to as the \textit{update locking problem} \cite{Jaderberg,Czarnecki}.
These considerations have motivated the development of alternative methods for credit assignment, each proposing solutions to some of these criticisms \cite{Lillicrap,Nokland,liao2016signsymmetry,frenkel2019learning,nokland2019localerror,clark2021credit,meulemans_tristany_2021}. However, none of the training schemes designed so far is able to comprehensively address \emph{all} the aforementioned challenges. 

Here we introduce a novel learning rule that is able to train ANNs on image classification tasks without incurring in the issues described above. We name our scheme PEPITA, Present the Error to Perturb the Input To modulate Activity. PEPITA relies on perturbing the input signal based on error-related information. The difference between the network responses to the input and to its perturbed version is used to compute the synaptic updates. Specifically, the algorithm is implemented by performing two forward computations for each input sample, avoiding any backward pass and performing the updates in a layerwise feedforward fashion during the second forward pass. By avoiding the backward pass, PEPITA does not suffer from the \textit{weight transport} problem and it partially solves the \textit{update locking} problem. Furthermore, as the error is incorporated in the input signal, PEPITA does not freeze the neural activity to propagate and apply the modulatory signal. Finally, the update rule respects the \textit{locality} constraints. We show that PEPITA can be formulated as a two-factor Hebbian-like learning rule. The error information is used as a global learning signal, which is consistent with biological observations of global 
neuromodulators that influence synaptic plasticity, and which is similar to several reinforcement learning schemes \cite{williams1992,Mazzoni}. As a proof-of-principle, 
we show that PEPITA can be successfully applied to train both fully connected and convolutional models, leading to performance only slightly worse than BP.

\section{Background and related work}
We review some of the training schemes proposed to solve the biologically unrealistic aspects of backpropagation. We begin by discussing learning rules relying on \textit{unit-specific feedback}, \textit{i.e.,} propagating the error from the output layer to each hidden layer through specific connectivity matrices. Then, we describe recent algorithms which involve delivering learning signals in alternative fashions.

\subsection{Credit assignment in conventional networks}
The backpropagation algorithm \cite{Rumelhart} relies on a forward and a backward pass. During the forward pass, the input signal is propagated from the input layer to the output layer, where the error is computed by comparing the network's output with the target. During the backward pass, the error flows from the top layers to the bottom layers through the same weights used in the feedforward pathway. For each synapse, the error gradient --- and the associated synaptic update --- is computed through recursive application of the chain rule. 
In the past two decades, several works investigated whether a mechanism similar to BP could be implemented by brains. \cite{Xie_Seung} showed the equivalence of BP and Contrastive Hebbian Learning, and, in a more general setting, \cite{scellier} introduced Equilibrium Propagation (EP), a learning framework for energy-based models, which shares similarities with Contrastive Hebbian Learning. EP plays a role analog to BP to compute the gradient of an objective function, and is a compelling biologically-plausible approach to compute error gradients in deep neuromorphic systems \cite{laborieux}. In the context of unsupervised learning, \cite{pehlevan} demonstrated that biologically plausible neural networks can be derived from similarity matching objectives, relying both on Hebbian feedforward and anti-Hebbian lateral connections.

Alternative training principles have been proposed to relax the constraint of symmetric weights. 
Effective learning in neural networks can be achieved also when the error is backpropagated through connections that share only the sign and not the magnitude with the feedforward weights (\textit{sign symmetry} algorithm) \cite{liao2016signsymmetry,xiao2018biologicallyplausible} or through random fixed connections (\textit{feedback-alignment} algorithm, FA) \cite{Lillicrap}. In the latter, the feedback provided by the random matrices is able to deliver useful modulatory information for learning since the forward connections are driven to align with the fixed feedback matrices. As an extension to FA, the \textit{direct feedback alignment} approach (DFA) shows that useful learning information can be propagated directly from the output layer to each hidden unit through random connectivity matrices \cite{Nokland,refinetti}. 
In a similar vein, \cite{Akrout2019DeepLW} builds on FA and introduces the \textit{weight mirror} neural circuit. This approach adjusts the initially random feedback weights of FA to improve their agreement with the forward connections, yielding improved performance compared to FA.

These biologically inspired training schemes achieve performance close to BP in many pattern recognition tasks without incurring in the \textit{weight transport} problem. However, they do not tackle the issues of non-locality, freezing of neural activity and update locking.

\subsection{Credit assignment without random feedback path}
Recently, learning techniques based on local error handling have been shown to successfully train ANNs while addressing the \textit{update locking} problem \cite{nokland2019localerror, belilovsky20,mostafa18}. In these approaches, however, each layer is trained independently through auxiliary fixed random classifiers, thereby incurring both in the \textit{weight transport} problem at the level of the classifiers and in a significant computational overhead. With the same goal, the \textit{direct random target projection} (DRTP) algorithm has been proposed to update the parameters of the hidden layers based on the sample labels rather than the network error \cite{frenkel2019learning}. Such a strategy overcomes both the \textit{weight transport} and the \textit{update locking} issues without incurring in additional computational requirements. However, it requires that the activity is frozen as the modulatory information on the targets flows through the network. Furthermore, DRTP suffers from a significantly larger performance degradation with respect to BP compared to the FA algorithms. 

Another original approach to credit assignment known as \textit{global error-vector broadcasting} (GEVB) avoids delivering error information through fixed random connections \cite{clark2021credit}. The GEVB learning rule performs parameter updates based on the inner product of the presynaptic activity and a global error vector. This scheme provides a performance almost on par with BP, and solves both the \textit{weight transport} and the \textit{locality} issues. However, it can only operate in a new class of deep neural networks, the vectorized nonnegative networks, which require each node to be represented by a vector unit with the same dimensionality as the output class. This implies a higher computational overhead, which significantly increases for datasets with a large number of classes, \textit{e.g.,} CIFAR-100.

\section{Error-driven input modulation}

\subsection{Overview of the proposed learning rule}

\begin{figure*}[h!]
     \centering
     \includegraphics[width=1\textwidth]{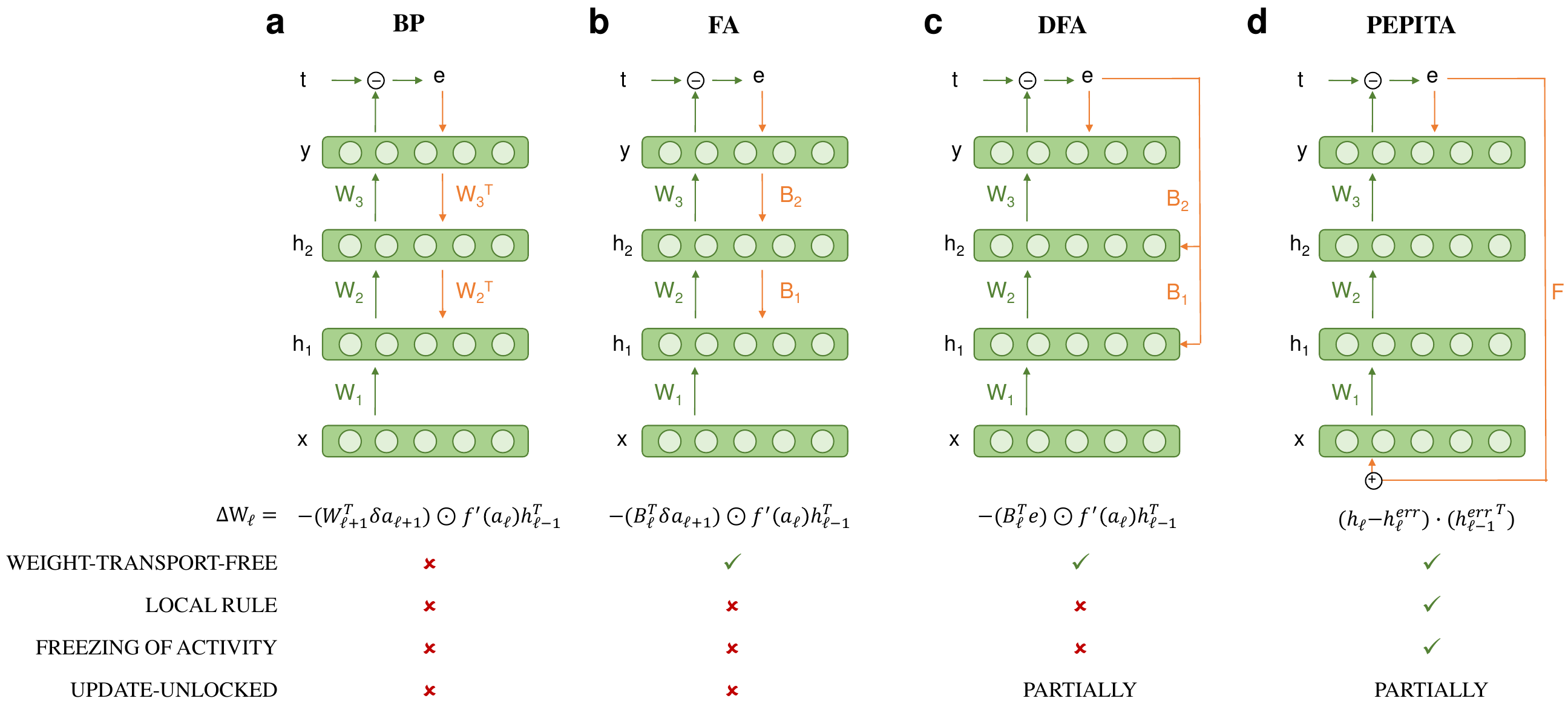}
     \vspace*{-5mm}
     \caption{Overview of different error transportation configurations. a) Back-propagation (BP). b) Feedback-alignment (FA). c) Direct feedback-alignment (DFA). d) Present the Error to Perturb the Input To modulate Activity (PEPITA). Green arrows indicate forward paths and orange arrows indicate error paths. Weights that are adapted during learning are denoted as $W_l$, and weights that are fixed and random are denoted as $B_l$ if specific to a layer (BP, FA, DFA) or $F$ if specific to the input signal (PEPITA). }
     \label{fig:schemes}
 \end{figure*}

Here we introduce a local plasticity rule to train ANNs via supervised learning. The training rules described so far rely on a forward pass, followed by a backward pass during which the error (BP, FA, DFA, GEVB) or the target (DRTP) travels from the output layer to each hidden unit through paths specific to each learning rule. The backward computation leads to at least two biologically implausible aspects. First, the weight updates rely on non-local information: the error coming directly from the output layer (all schemes), as well as downstream feedback connections (BP, FA). Second, during the backward pass the network activity is frozen. We circumvent both problems by proposing a training scheme that does not involve a backward pass, but rather performs two successive forward passes per input sample. The first pass is similar to any conventional training scheme. In the second forward pass, the error computed during the first pass is used to modify the input. The modulated input travels through the network, eliciting activities slightly different from those of the first pass. Such differences in node activity are used to update the network's parameters. As the output and input dimensionalities are generally different, we use a fixed random matrix F, with zero mean and small standard deviation, to project (\textit{i.e.,} add) the error onto the input. We refer to the first and second forward passes as ``standard pass'' and ``modulated pass'', respectively. Figure \ref{fig:schemes} shows a schematic comparison between  error propagation in BP, FA and DFA and the novel configuration proposed in this paper, PEPITA. The intuition of modulating the input signal with error-related information is motivated by the existence of global modulatory  signals in the brain fed from higher level to lower level connections that influence the activity at the early stages of the visual stream \cite{Shimegi2016CholinergicAS}.

\subsection{The learning rule}\label{sec:learning_rule}
Given a fully connected neural network with $L$ layers and an input signal $x$, in the standard pass the hidden unit and output unit activations are computed as:

\begin{equation}\label{eq:standardpass}
\begin{split}    
    h_1 = \sigma_1 (W_1 x), \\
    h_l = \sigma_l (W_l h_{l-1}) \hspace{0.3cm} 2\leq l \leq L.
\end{split}
\end{equation}

where $\sigma_l$ denotes the non-linearity at the output of the $l^{th}$ layer and $W_l$ denotes the matrix of weights between layer $l-1$ and layer $l$. At the modulated pass, the activations are computed as:
\begin{equation}\label{eq:modulatedpass}
\begin{split}
    h_1^{err} = \sigma_1 (W_1 (x + F e) ), \\
    h_l^{err} = \sigma_l (W_l h_{l-1}^{err})  \hspace{0.3cm} 2\leq l \leq L.
\end{split}
\end{equation}

where $e$ denotes the network error and $F$ denotes the fixed random matrix used to project the error on the input. We compute the error as $e = h_L - target$.\\
During the two forward passes, the network weights are equal and the difference in the node activities is solely due to the error-driven modulation of the input signal. \\
The weight updates are computed after (or during, see Discussion) the modulated pass and are applied in a forward fashion. Each synaptic update depends on the product between a postsynaptic term and a presynaptic term. The postsynaptic term is given by the difference in activation of the postsynaptic node between the standard and the modulated pass. The presynaptic term corresponds to the activity of the presynaptic nodes during the modulated pass. 

First layer: \\ 
\begin{equation}
\Delta W_1 = (h_1-h_1^{err})\cdot (x + Fe)^T 
\end{equation}

Intermediate hidden layers, with $2\leq l\leq L-1$: 
\begin{equation}
\Delta W_l = (h_l-h_l^{err})\cdot (h_{l-1}^{err})^T 
\end{equation}

Output layer:  
\begin{equation}
\Delta W_L = e\cdot (h_{L-1}^{err})^T 
\end{equation}
The weights of the output layer are trained as in BP since the information about the error is directly accessible at the last layer. Furthermore, we tested a modification of the algorithm and replaced the activity of the modulated pass with that of the standard pass in the presynaptic term: $\Delta W_l = (h_l-h_l^{err})\cdot (h_{l-1})^T $. In the simulations, such modification did not affect the network performance.

Finally, the prescribed synaptic updates are applied depending on the chosen optimizer, as for any gradient-based optimization technique. For example, using the Stochastic Gradient Descent (SGD)-like scheme, with learning rate $\eta$:
\begin{equation}\label{eq:SGD}
    W(t+1) = W(t) - \eta \Delta W
\end{equation}

The algorithm is reviewed in the pseudocode below.

\begin{algorithm}
\caption{Implementation of PEPITA}\label{pseudocode}
\textbf{Given:} Input (\textit{x}) and label (\textit{target})\\
\color{gray}\#standard forward pass \color{black}\\
\textit{$h_0$ = x} \\
\textbf{for} $\ell$ = 1, ..., L\\
\hphantom{xxx} $h_{\ell} = \sigma_\ell(W_\ell h_{\ell-1}) $ \\
$e = h_L -$ \textit{target} \\
\color{gray}\#modulated forward pass \color{black}\\
\textit{$h_0^{err}$ = x+Fe} \\
\textbf{for} $\ell$ = 1, ..., L\\
\hphantom{xxx} $h_{\ell}^{err} = \sigma_\ell(W_\ell h_{\ell-1}^{err}) $ \\
\hphantom{xxx} \textbf{if} $\ell < L$:\\
\hphantom{xxx}\hphantom{xxx} $\Delta W_\ell = (h_\ell-h_\ell^{err})\cdot (h_{\ell-1}^{err})^T $\\
\hphantom{xxx} \textbf{else:}\\
\hphantom{xxx}\hphantom{xxx} $\Delta W_\ell = e\cdot (h_{\ell-1}^{err})^T $
\end{algorithm}


 \begin{figure*}[h!]
     \centering
     \includegraphics[width=1\textwidth]{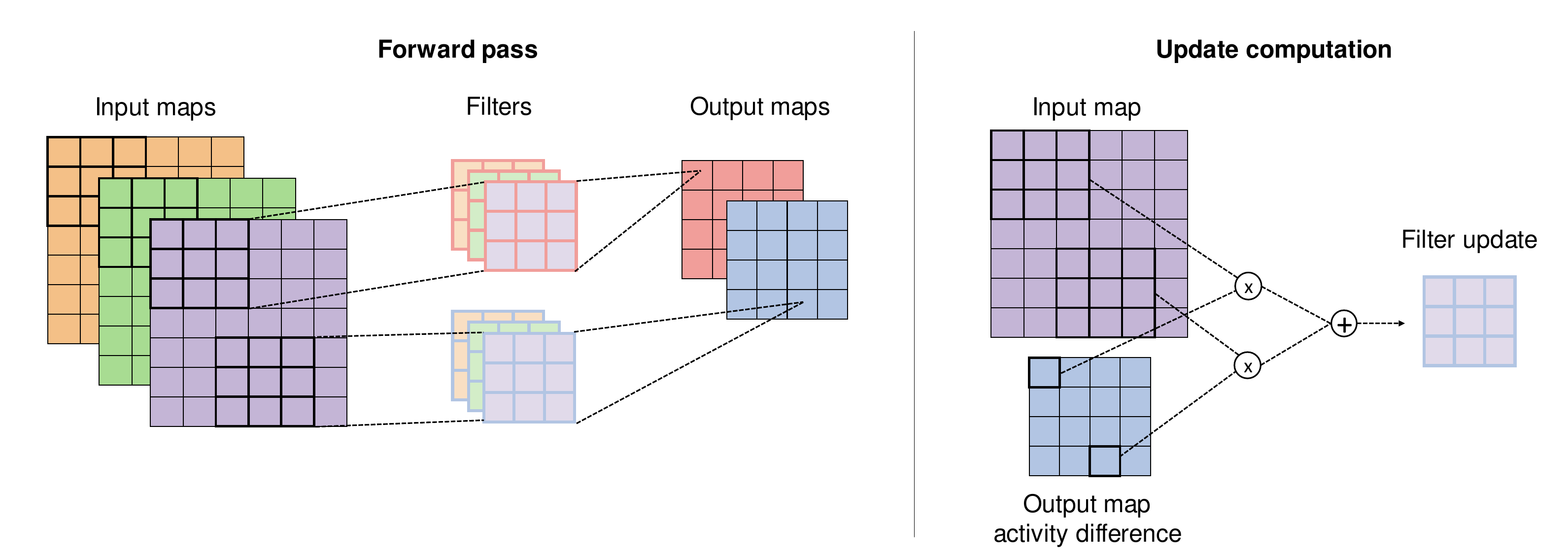}
     \vspace*{-8mm}
     \caption{Update scheme for convolutional layers}
     \label{fig:conv}
 \end{figure*}

\subsection{Extension to convolutional layers}
The same approach can be applied to train convolutional models. However, a modification is required to account for the parameter sharing used in convolutions. As for fully connected models, PEPITA trains the convolutional models through a standard and a modulated forward pass, and the kernels are learnt based on post- and pre-synaptic related terms. For each filter, the update is computed after the modulated pass through the following steps:
\begin{enumerate}
    \item For each element in the output map, first we compute the difference in activity between the standard and modulated pass. Then we multiply such difference by the area of the input map on which the filter was applied to generate the specific output pixel. Each of the computed products has the same dimensionality as the filter.
    \item All the products computed in step 1 are summed into a single update term.
    \item The resulting term is divided by the number of summed products.
    \item The computed update is applied to the filter with the chosen optimization technique, such as Eq. \ref{eq:SGD}.
\end{enumerate}
Figure \ref{fig:conv} provides a schematic of the standard forward pass and the computation of the update for a single channel of a filter.

\section{Results}
\subsection{Methods}
Inspired by \cite{liao2016signsymmetry}, we focus on relative differences between algorithms, not absolute performance. Each experiment is a \{model, dataset\} pair. We tested the PEPITA learning rule on 3 datasets: MNIST \cite{MNIST}, CIFAR-10 \cite{CIFAR10krizhevsky} and CIFAR-100 \cite{CIFAR100}. We did not use any data augmentation. For each dataset, we trained both a fully connected and a convolutional model with BP, FA, DRTP and PEPITA. 
With BP, FA and PEPITA, we used rectified linear unit (ReLU) \cite{Nair_relu} as non-linearity for the hidden layers and softmax for the output layer. With DRTP we used hyperbolic tangent as non-linearity for the hidden layers and sigmoid for the output layer. We used SGD with momentum with hyperparameter 0.9. For BP, FA and DRTP we used cross entropy loss. In the fully connected networks, we introduced dropout with drop rate of 10\% after the hidden layer. In the convolutional models, we applied Max Pooling after the convolutional layer. We initialized the networks using the He initialization \cite{he_init} for both the forward connections $W_i$ and the projection matrix $F$. The input images are normalized to the interval [0,1]. We optimized the learning rate, the learning rate decay schedule, the batch size and the dropout rate separately for each experiment. We used the entire training set for the training and did not use a validation set. The network architectures and simulation details for each dataset and learning rule are reported in Appendix \ref{app:sim_details}.

The code for the simulations of fully connected models with PEPITA, BP, FA and convolutional models with PEPITA is provided at: \url{https://github.com/GiorgiaD/PEPITA} . For all the other experiments we used the code from \cite{frenkel2019learning}.

\subsection{Experimental results}
\begin{figure}[h!]
     \centering
     \includegraphics[width=0.5\textwidth]{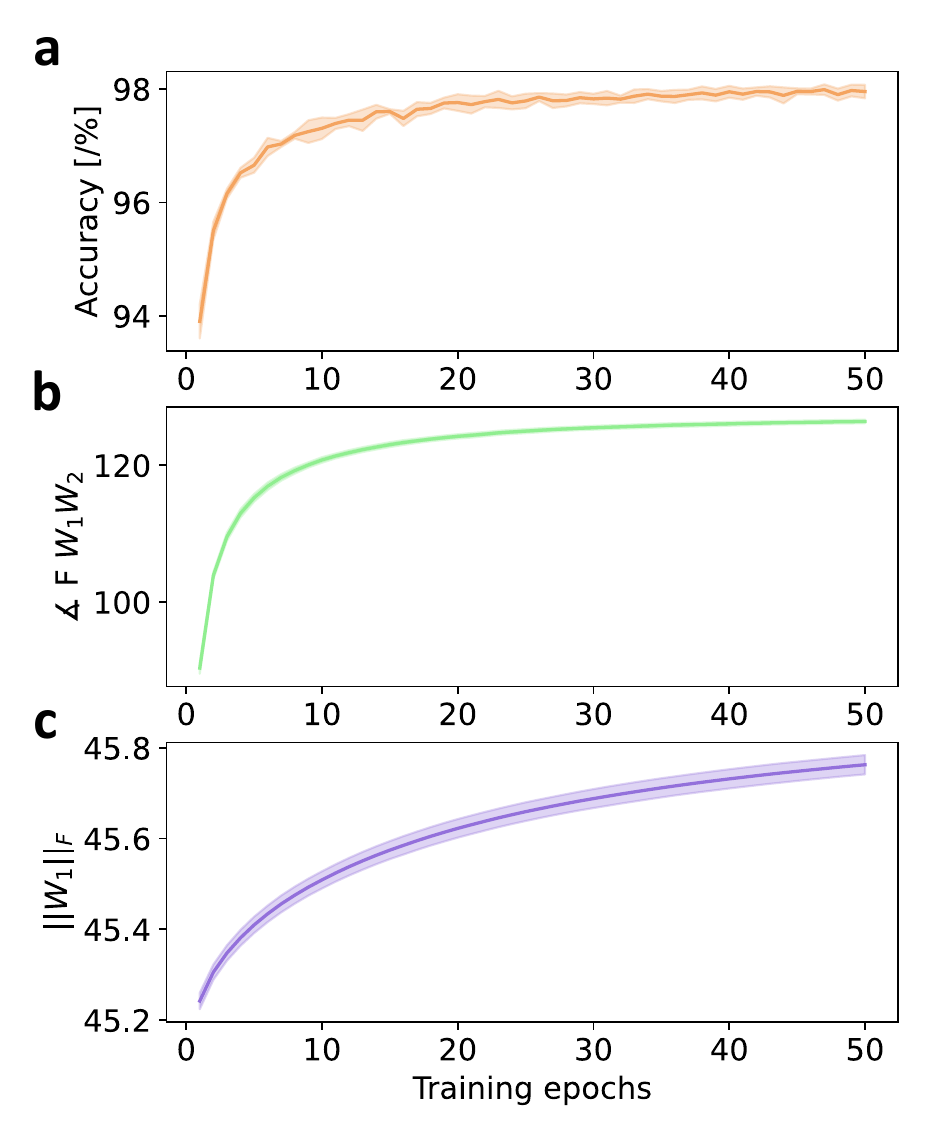}
     \vspace*{-8mm}
     \caption{Dynamics of a 1 hidden layer network, with 1024 hidden units, 10\% dropout, during training on MNIST with PEPITA. (a) Test accuracy computed after each training epoch. (b) Angle between the product of the forward matrices and the F matrix. (c) Evolution of the norm of the initial weight matrix. The solid line reports the mean over five independent runs, the shaded area indicates the standard deviation.}
     \label{fig:training}
 \end{figure}

\begin{figure}[h!]
     \centering
     \includegraphics[width=0.5\textwidth]{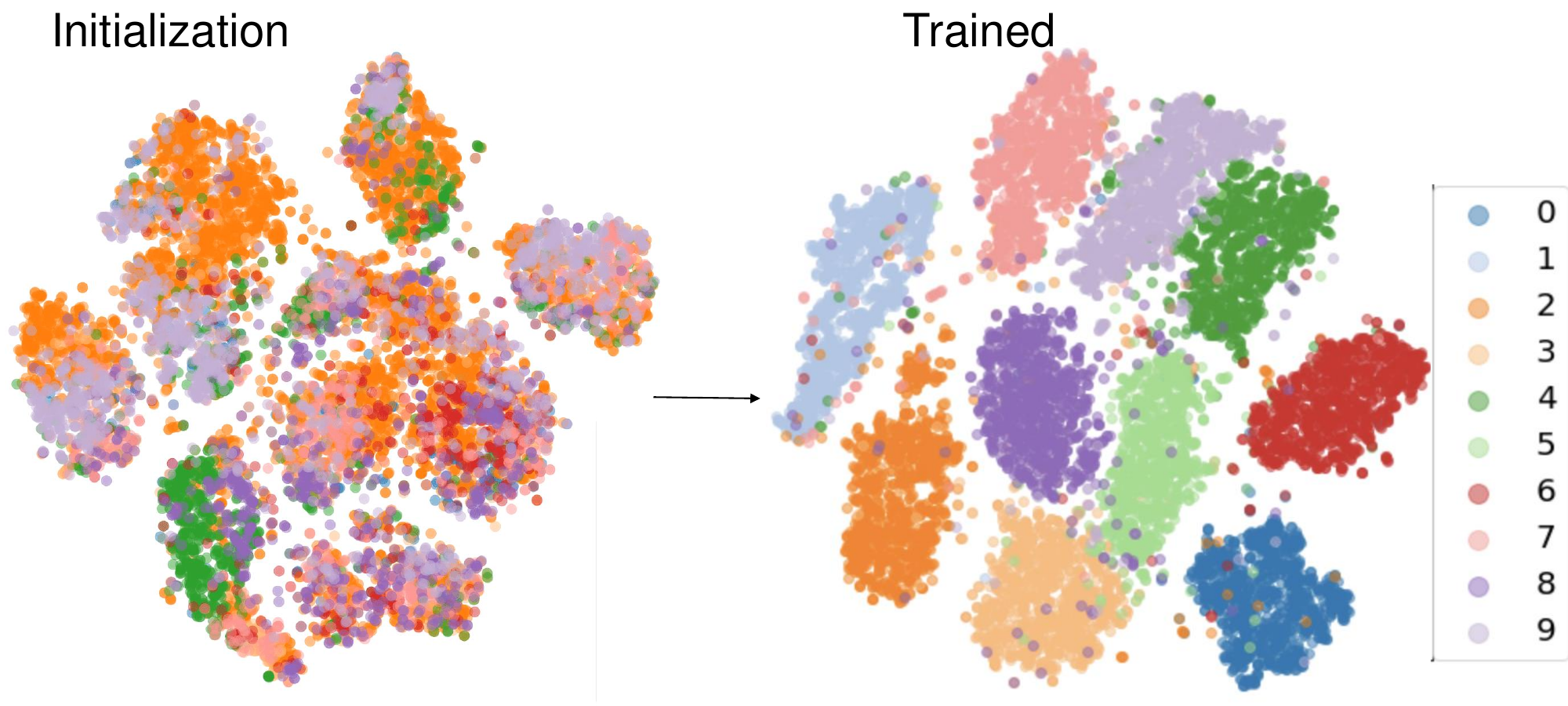}
     \vspace*{-5mm}
     \caption{t-SNE embeddings of the MNIST representation at the hidden layer of the fully connected model trained on MNIST, both before (initialization) and after training (trained). Each color corresponds to a different class. }
     \label{fig:clustering}
 \end{figure}

Here, we show that the proposed learning rule successfully trains both fully connected and convolutional networks on image classification tasks. The experimental results are summarized in Table \ref{tab:results}. The accuracy obtained by PEPITA for fully connected networks (Table \ref{tab:results}, columns 1-3) is close to the accuracy achieved by BP and FA, and superior to that of DRTP for all experiments. Figure \ref{fig:training}a shows the learning curve on the test set for the fully connected model trained on MNIST with PEPITA. Figure \ref{fig:clustering} reports the t-SNE visualization of the representation learned at the hidden layer for the same model. Appendix \ref{app:test_curves} reports PEPITA's test curve on the CIFAR-10 dataset, compared with the test curves of BP, FA, and DRTP. Figure \ref{fig:training}a and Figure \ref{fig:comparison} show that the training rules lead to convergence within 100 epochs.
 We observe an improved performance of the convolutional models trained with PEPITA compared to the fully connected networks for all datasets (Table \ref{tab:results}, columns 4-6), indicating that the convolutional version of PEPITA is able to learn useful two-dimensional filters. 

To evaluate the learning speed of the tested training schemes, we relied on the \textit{plateau equation for learning curves} proposed in \cite{Dellaferrera_GRAPES}:
\vspace{-1.5mm}
\begin{equation}
    \textnormal{accuracy} = \frac{\textnormal{max\_accuracy} \cdot \textnormal{epochs}}{\textnormal{slowness} + \textnormal{epochs}}.
    \vspace{-2mm}
\end{equation}

By fitting the test curve to this function, we extract the \textit{slowness} parameter, which quantifies how fast the network reduces the error during training. In Appendix \ref{app:slowness} we show the \textit{plateau curve} fitting the learning curves of BP, FA, DRTP, and PEPITA for the fully connected model trained on CIFAR-10. The table in Appendix \ref{app:slowness} 
reports the slowness values for the fully-connected models on all datasets. PEPITA's convergence rate is in between BP (the fastest) and FA (the slowest). DRTP has a better learning speed than FA, but it converges to a significantly lower accuracy plateau.

\begin{figure}[h!]
 \centering
 \includegraphics[width=0.45\textwidth]{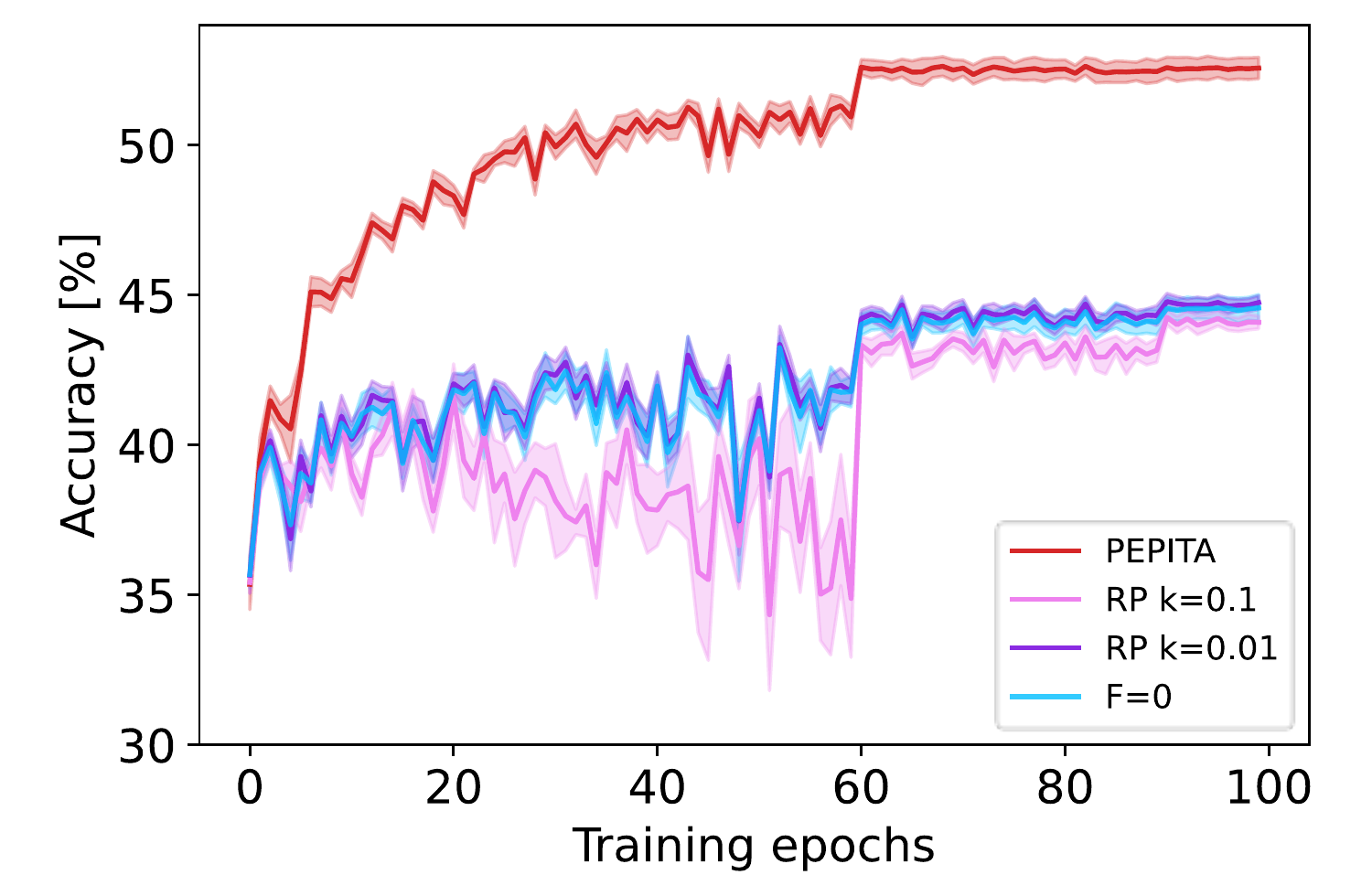}
 \vspace*{-3mm}
 \caption{Test curve for PEPITA in its original formulation, PEPITA with F=0 (\textit{i.e., }only the last layer is trained), and a Random Projection (RP) algorithm in which the input is modulated with a random vector $\textit{r}$. The elements of $r$ are sampled from a uniform distribution over [-1, 1) and multiplied by a scalar $k$. We used the same settings as in the experiments for the fully connected models trained on CIFAR-10 (second column of Table \ref{tab:results}). The solid line is the mean over 10 independent runs. The shaded colored area shows the std over the 10 runs. PEPITA achieves a significantly higher accuracy than the F=0 and RP schemes. Additionally, for small $k$ values, the RP algorithm (purple) has a similar learning curve as the F=0 algorithm (blue).}
 \label{fig:F0&RP_main}
\end{figure}

In PEPITA, the error-based perturbation of the input is key to directing the parameter updates in the correct direction. We verify that the accuracy is above chance but much lower than with PEPITA, if in the second forward pass the input is perturbed with random noise, or if F is set to zero (\textit{i.e.}, only the last layer is updated in a BP-fashion). Figure \ref{fig:F0&RP_main} compares the test curves obtained on CIFAR-10 with PEPITA, using F=0 or random noise modulation with different amplitudes. Our results show that, if the input is not perturbed or is perturbed independently from the error, the network exhibits worse performance. Additional details on the simulations are reported in Appendix \ref{app:F0&RP}.

To gain insight into the dynamics of the PEPITA-based learning rule, we considered a two-layer network and measured the alignment of the product between the forward weights ($W_1 W_2$) with the fixed matrix $F$. As in \cite{xiao2018biologicallyplausible}, we flattened the matrices into vectors and computed the angle between the vectors. Figure \ref{fig:training}b reports the alignment dynamics of a network with one hidden layer during training on MNIST. We observe that, at initialization, the alignment angle is $\sim$90$^{\circ}$ (\textit{i.e.,} random), and that during training the angle increases, saturating at approximately $120^{\circ}$. The evolution of the angle alignment finds a plateau as the test accuracy saturates. Interestingly, we observe that our approach encourages a soft `antialignment' (\textit{i.e.,} the angle increases above $90^{\circ}$) of the forward matrices with the fixed feedback matrix.
We provide an intuitive explanation of this phenomenon in section \ref{sec:analytic}. 

\begin{table*}[t]
\caption{Test accuracy [\%] achieved by BP, FA, DRTP and PEPITA in the experiments. Mean and standard deviation are computed over 10 independent runs.}
\label{tab:results}
\vskip 0.15in
\begin{center}
\begin{small}
\begin{sc}
\begin{tabular}{l|ccc|ccc}
\toprule
& \multicolumn{3}{|c|}{Fully connected models} & \multicolumn{3}{|c}{Convolutional models} \\ \midrule
  & MNIST & CIFAR10 & CIFAR100 & MNIST & CIFAR10 & CIFAR100 \\
\midrule
 BP & 98.63$\pm$0.03 & 55.27$\pm$0.32 & 27.58$\pm$0.09 & 98.86$\pm$0.04 &  64.99$\pm$0.32  & 34.20$\pm$0.20 \\ 
 FA & 98.42$\pm$0.07 & 53.82$\pm$0.24 & 24.61$\pm$0.28 & 98.50$\pm$0.06 & 57.51$\pm$0.57 & 27.15$\pm$0.53 \\ 
 DRTP & 95.10$\pm$0.10 & 45.89$\pm$0.16 & 18.32$\pm$0.18 & 97.32$\pm$0.25 & 50.53$\pm$0.81 & 20.14$\pm$0.68 \\
 PEPITA & 98.01$\pm$0.09 & 52.57$\pm$0.36 & 24.91$\pm$0.22 & 98.29$\pm$0.13 & 56.33$\pm$1.35 & 27.56$\pm$0.60 \\ 
\bottomrule
\end{tabular}
\end{sc}
\end{small}
\end{center}
\vskip -0.1in
\end{table*}

\begin{figure*}[h!]
 \centering
 \includegraphics[width=1\textwidth]{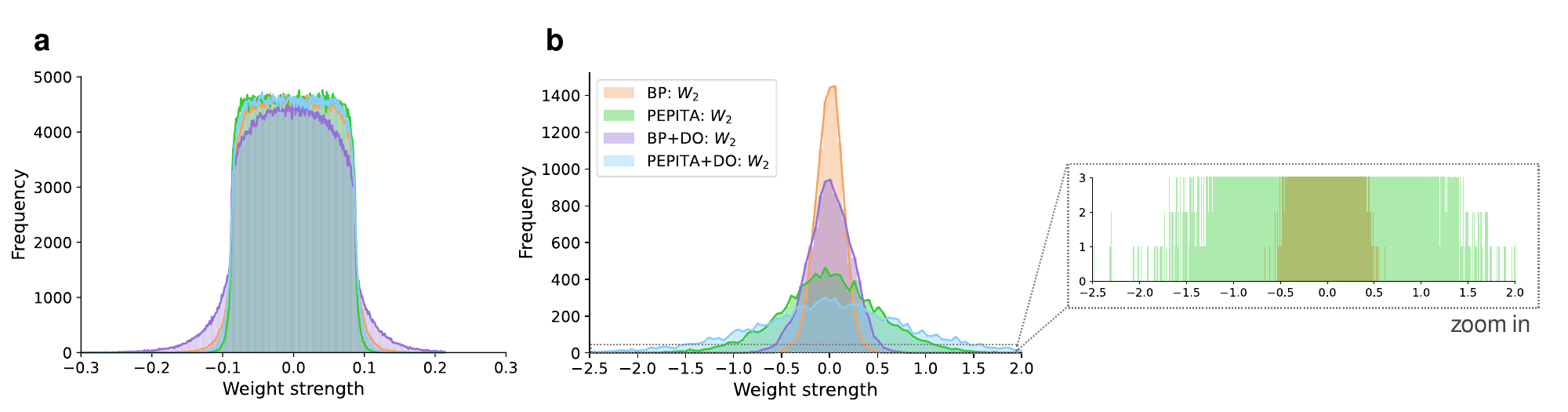}
 \vspace*{-7mm}
 \caption{Distribution of the weights trained with BP (orange, purple) and PEPITA (green, blue) on MNIST for a network with one hidden layer with 1024 units, with and without dropout (DO), with dropout rate 10\%. (a) First hidden layer (b) Output layer }
 \label{fig:weights}
\end{figure*}

\subsection{Analytic results}\label{sec:analytic}
We observed experimentally that the intrinsic dynamics of PEPITA tend to drive the product between the forward weight matrices to `antialign' with the $F^T$ matrix (Figure \ref{fig:training}b). In Appendix \ref{app:proof} we provide a formal proof showing why this phenomenon occurs in the case of a linear fully connected network. Such analytical result for a linear network supports our empirical finding on non-linear models that the product of the forward matrices comes to antialign with $F^T$. We conjecture that such antialignment enables the error that modulates the input, and consequently the neural activity, to deliver useful learning signals for weight updates in $A$ and $W$. We remark that with a different sign choice for the input modulation step in Eq. \ref{eq:modulatedpass} -- \textit{i.e., $h_1^{err} = \sigma_1 (W_1 (x - F e) )$} -- the product of the forward matrices would instead come to `align' with $F^T$. In the proof, we make the assumption that the norm of the first hidden layer weights increases monotonically. We verify empirically that this is a realistic assumption in Figure \ref{fig:training}c. Appendix \ref{app:wnorm} shows that the rate of weights’ norm increase diminishes as training continues beyond 50 epochs and the learning rate is decayed. Furthermore, the proof relies on the assumption that the input is whitened. In Appendix \ref{app:whiten} we show that PEPITA effectively trains networks also with whitened input.


\subsection{Final weight distribution}
We compared the layer-wise weight distribution of networks trained with BP and PEPITA, initialized with the same normal distribution \cite{he_init}. For a two-layer model trained on MNIST, the trained weights in the first layer have similar distributions for BP and PEPITA (Figure \ref{fig:weights}a), with the weight distribution for BP covering a slightly greater range than for PEPITA. The distribution in the last layer is instead significantly different: both BP and PEPITA lead to bell-shaped distributions with mean close to zero, however the distribution width obtained with PEPITA is significantly larger (Figure \ref{fig:weights}b). Importantly, this implies that PEPITA learns different solutions with respect to BP, and those solutions yield successful performance. In Appendix \ref{app:weights} we show that the weight distribution obtained with BP can be fit with a Normal curve, while the distribution obtained with PEPITA is better described by a Student's t-distribution. The Student's t-distribution is a subexponential distribution, which is reminiscent of the connectivity structure of neurons in cortical networks, where the amplitudes of excitatory post-synaptic potentials obey a long-tailed pattern, typically lognormal \cite{buszaki,song, Iyer,Wang2021}. 
Furthermore, PEPITA drives the training towards a parameter distribution in which most synaptic connections are weak and a few weights are sparse and strong (zoom of Figure \ref{fig:weights}b). 
\cite{valiant09} demonstrates that a network regime featuring some maximally strong synapses enables sparse and disjoint representation of items. This yields a significantly higher memory capacity compared to a regime supported by weak synapses only. 
We also remark that previous work \cite{blundell} has shown that networks regularized with dropout \cite{hinton_dropout} or uncertainty on the weights (\textit{i.e.,} Bayes by Backprop \cite{blundell}) present a greater range of weight strength as compared to standard SGD. 
Therefore, PEPITA, which is conceived to solve the biological implausible aspects of network training via BP, conveys the network weights resembling more biologically plausible distributions than BP, possibly supporting higher memory capacities as well as more regularized parameters. However, we observe that, despite the similarity in subexponential distribution, the weight distribution learnt by PEPITA differs from the synaptic strength in biological circuits in the ratio between excitatory to inhibitory connections. While it is generally thought that biological neurons function under balanced excitation and inhibition, this does not imply equal numbers of excitatory and inhibitory neurons as it occurs in PEPITA.

\section{Discussion}
We introduced PEPITA, a learning rule that relies on perturbing the input signal through error-related information and updating the network parameters based on local variations of activity. Our results indicate that PEPITA is able to train both fully connected and convolutional neural networks on image classification tasks, with an accuracy comparable to the feedback alignment algorithms. We empirically observe that PEPITA encourages the feedforward weights to evolve towards a soft `antialignment' with the feedback matrix, and it nudges the weight distribution towards a ``strong-sparse and weak-dense'' distribution reminiscent of the connectivity of cortical neurons. 

By replacing the backward pass with an additional forward pass, our approach avoids BP's biologically unrealistic aspects of weight transport, non-locality, freezing of neural activity during parameter updates, and update locking. First, PEPITA is not affected by the weight transport problem as it does not send weight-specific feedback signals. Secondly, regarding locality, the prescribed updates of each synapse solely rely on the activity of the nodes it is connected to. The only global signal required by the algorithm is the error projected onto the input. The role of this global error in perturbing the node activity may correspond to a neuromodulator that influences local synaptic plasticity \cite{Mazzoni,williams1992,clark2021credit}. Third, during training, PEPITA never requires the network activity to be frozen since the error is propagated together with the input signal. Finally, in PEPITA the updates are performed layer-wise in a feedforward fashion allowing to partially solve the update locking issue. Indeed, the weights of the first layer can be updated right at the beginning of the second forward pass and do not need to wait for the updates of the downstream layers. Hence, the forward computation for the next input sample at the first layer can start in parallel with the update of the second layer from the previous sample, and so on. In the case of very deep networks, such a strategy could substantially reduce the computational time, suggesting that PEPITA could be suitable for edge computing devices requiring fast processing of the input signals.

While our approach addresses BP's issues of biological plausibility, it introduces additional elements with respect to BP and FA. Specifically, PEPITA involves the projection of the error onto the input through a fixed random matrix. We speculate that the role of such a projection matrix is reminiscent of the connectivity matrices supporting the cortico-thalamo-cortical loops throughout neocortex. Additionally, the error-based input modulation is supported by the observation that neuromodulators can modulate the activity in the visual stream. Although neuromodulators do not change activity in the retina (let alone inputs to the retina), they can change the activity at the very early stages of the visual pathway (LGN, V1) \cite{Kreiman2021}. Since we apply PEPITA on shallow networks, the algorithm approximates the role of neuromodulators by changing the input directly. In deeper networks, we could apply the error-based modulation on the low hidden layers rather than on the input.  
Furthermore, PEPITA requires storing the activations of the standard pass during the modulated pass. We suggest that this could be implemented in biological neurons through dendritic and somatic activities. Earlier works have proposed two-compartmental neuron models in which learning is driven by local prediction mismatch between synaptically-driven dendritic activity and somatic output \cite{asabuki20,sacramento18}. We propose that if we apply PEPITA to train a network of such two-compartment neurons, the response to the  input signal in the standard forward pass is first integrated in the dendritic compartment and then encoded in the somatic activity. Then, in the modulated forward pass, the dendrites encode the response to the modulated input while the soma is still storing the response to the original input. The mismatch between the dendritic activity encoding the response to the modulated input and the somatic activity encoding the response to the original input drives the synaptic update. 
Consequently, synaptic strength can be updated based on the outcome of the second pass and the outcome of the first pass through a mismatch between somatic activity (storing the activity of the first pass) and dendritic activity (integrating the activity of the second pass) in two-compartment neurons. Alternatively, the activity of the first pass can drive plasticity through the dynamics of interneurons: the past activation can be stored in an interneuron and then retrieved after the second forward pass.

Importantly, PEPITA bears structural similarity with learning rules that are believed to take place in the brain, \textit{i.e.,} Hebbian-like approaches. Indeed, the update rule shown in the pseudocode 
can be decoupled into two separate successive updates: $\Delta W_\ell^1 = (h_\ell)\cdot (h_{\ell-1}^{err})^T $ and $\Delta W_\ell^2 = (-h_\ell^{err})\cdot (h_{\ell-1}^{err})^T $.
Each of the two updates embodies a fundamental feature of Hebbian learning \cite{gerstner_neuronal_dynamics}, \textit{i.e.,} the synaptic strength modification is proportional to both a presynaptic and a postsynaptic signal. Therefore the definition of PEPITA fits in the criteria to be a biologically plausible mechanism. 

In conclusion, we have demonstrated that a learning rule that solves the biologically implausible aspects of BP by relying only on forward computations, is able to train both fully connected and convolutional models with a performance close to BP and on par with FA. The proposed algorithm can thus help bridge the gap between neurobiology and machine learning.

\section*{Acknowledgements}
We thank the reviewers for useful comments to the paper. We are grateful to W. Xiao, T. Bricken, G. Indiveri, L. Petrini, A. Cooper, M. Pizzochero, and our colleagues at the Kreiman Lab for fruitful discussion. This work was supported by NIH grant R01EY026025 and NSF grant CCF-1231216.

\bibliography{example_paper}
\bibliographystyle{icml2022}

\newpage
\appendix
\onecolumn

\section{Simulation details.}\label{app:sim_details}

\begin{table*}[h!]
\footnotesize
\label{tab:architectures}
\vskip 0.15in
\begin{center}
\begin{small}
\begin{sc}
\begin{tabular}{l|lll|lll}
\toprule
& \multicolumn{3}{|c|}{Fully connected models} & \multicolumn{3}{|c}{Convolutional models} \\ \midrule
  & MNIST & CIFAR10 & CIFAR100 & MNIST & CIFAR10 & CIFAR100 \\
\midrule
InputSize & 28$\times$28$\times$1 & 32$\times$32$\times$3 & 32$\times$32$\times$3 & 28$\times$28$\times$1 & 32$\times$32$\times$3 & 32$\times$32$\times$3 \\ 
 Layer 1 & FC1:1024 & FC1:1024 & FC1:1024 & Conv1:32,5,1 & Conv1:32,5,1 & Conv1:32,5,1 \\ 
 Layer 2 & FC2:10 & FC2:10 & FC2:100 & FC1:10 & FC1:10 & FC1:100 \\ \midrule
 $\eta$ BP & 0.1 & 0.01 & 0.01 & 0.1 & 0.01 & 0.01 \\ 
 $\eta$ FA & 0.1 & 0.001 & 0.001 & 0.1 & 0.01 & 0.01\\ 
 $\eta$ DRTP & 0.01 & 0.001 & 0.001 & 0.01 & 0.01 & 0.01 \\ 
 $\eta$ PEPITA & 0.1 & 0.01  & 0.01 & 0.1 & 0.1 & 0.1 \\ 
 $\eta$ decay rate & $\times$0.1 & $\times$0.1 & $\times$0.1 & $\times$0.1 & $\times$0.1  & $\times$0.1 \\
 decay epoch & 60 & 60,90 & 60,90 & 10,30,50 & 10,30,50  & 10,30,50  \\
 Batch size & 64 & 64 & 64 & 100 & 100 & 100 \\ 
 F std & 0.05$\cdot 2\sqrt{\frac{6}{\text{fanin}}}$ & 0.05$\cdot 2\sqrt{\frac{6}{\text{fanin}}}$ & 0.05$\cdot 2\sqrt{\frac{6}{\text{fanin}}}$ & 0.05$\cdot 2\sqrt{\frac{6}{\text{fanin}}}$ & 0.05$\cdot 2\sqrt{\frac{6}{\text{fanin}}}$ & 0.05$\cdot 2\sqrt{\frac{6}{\text{fanin}}}$ \\ 
 Fan in & $28\cdot28\cdot1$ & $32\cdot32\cdot3$ &  $32\cdot32\cdot3$ & $28\cdot28\cdot1$ & $32\cdot32\cdot3$ &  $32\cdot32\cdot3$ \\ 
 $\#$epochs & 100 & 100 & 100 & 100 & 100 & 100 \\
 Dropout & 10\% & 10\% & 10\% & - & - & - \\ 
 Weight init & He unif. & He unif. & He unif. & He unif. & He unif. & He unif. \\ 
\bottomrule
\end{tabular}
\end{sc}
\end{small}
\end{center}
\caption{Network architectures and settings used in the experiments. For the convolutional layers A,B,C means A feature maps of size B, with stride C.}
\vskip -0.1in
\end{table*}

\newpage
\section{Test curves}\label{app:test_curves}
\begin{figure*}[h!]
 \centering
 \includegraphics[width=0.6\textwidth]{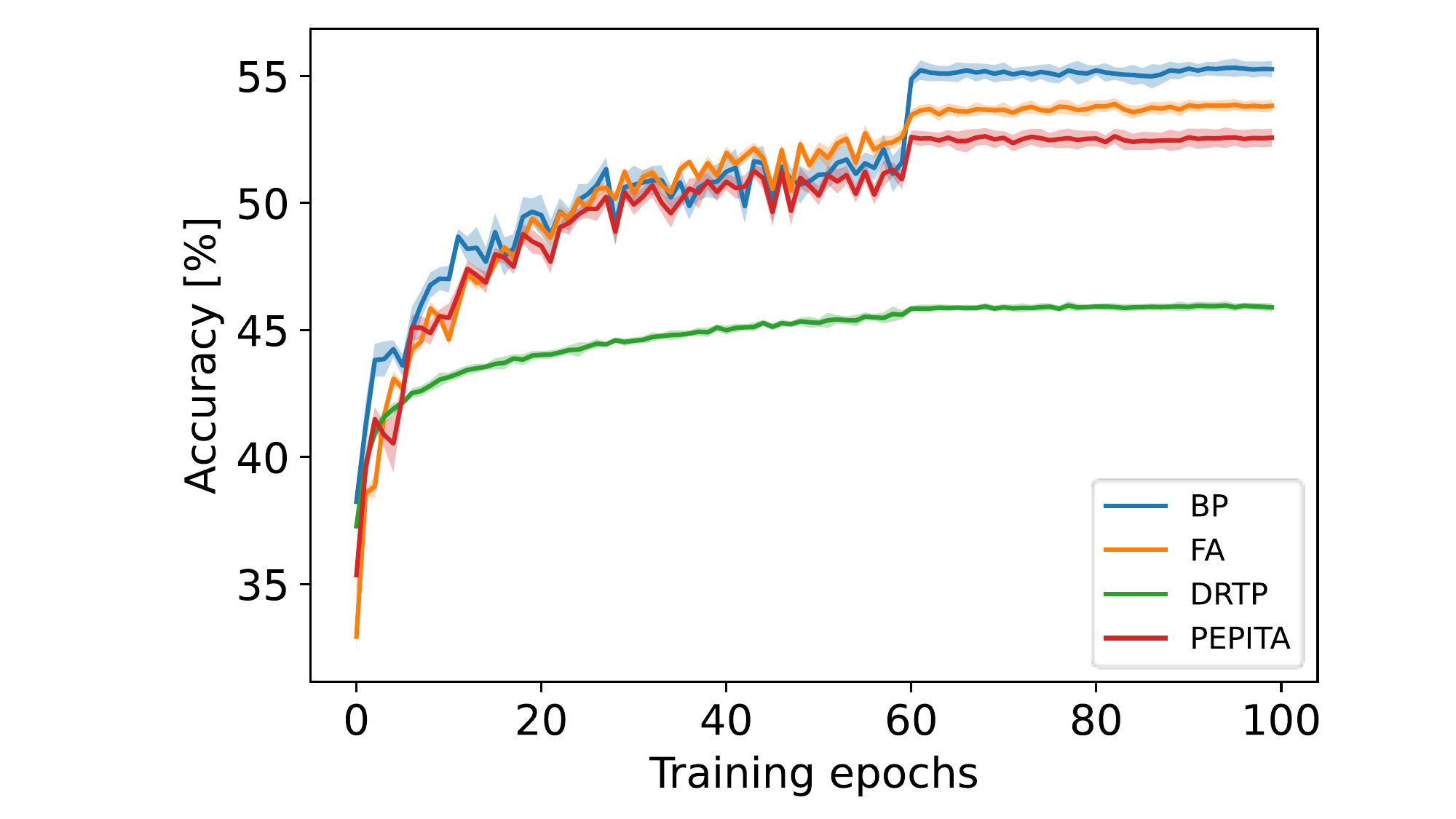}
 \caption{Test curve for BP, FA, DRTP and PEPITA in the experiments for the fully connected models trained on CIFAR-10 (second column of Table \ref{tab:results}). The solid line is the mean over 10 independent runs. The shaded colored area shows the std over the 10 runs. }
 \label{fig:comparison}
\end{figure*}

\newpage
\section{Learning speed}\label{app:slowness}
\begin{figure*}[h!]
 \centering
 \includegraphics[width=0.85\textwidth]{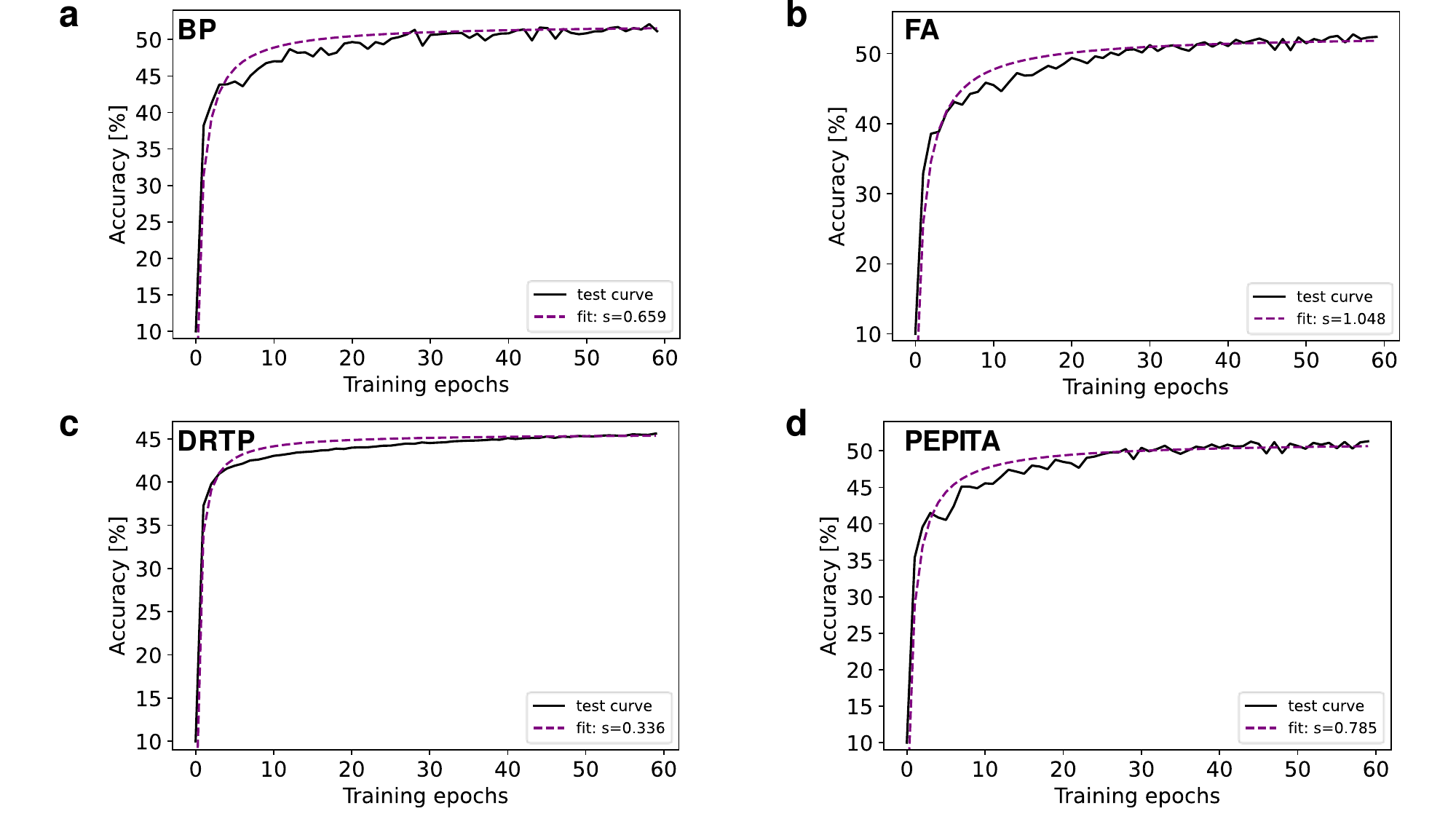}
 \caption{The \textit{plateau equation for learning curves} fits the test curves. Results for the fully connected models trained on CIFAR-10 (second column of Table \ref{tab:results}). The black solid line is the test curve averaged over 10 independent runs. The purple dotted line is the fitted curve from which we extract the slowness value (s). Results for (a) BP, (b) FA, (c) DRTP, and (d) PEPITA. }
 \label{fig:slowness}
\end{figure*}

To evaluate the convergence rate, we relied on the \textit{plateau equation for learning curves} proposed in \cite{Dellaferrera_GRAPES}:
\begin{equation}
    \textnormal{accuracy} = \frac{\textnormal{max\_accuracy} \cdot \textnormal{epochs}}{\textnormal{slowness} + \textnormal{epochs}}.
\end{equation}
By fitting the test curve to this function, we extract the \textit{slowness} parameter, which quantifies how fast the network reduces the error during training. Mathematically, the \textit{slowness} value corresponds to the number of epochs necessary to reach half of the maximum accuracy. Hence, the lower the \textit{slowness}, the faster the training. In our simulations, we perform the fit on the first 60 epochs, as on epoch 61 the learning rate is decayed and the test curve exhibits a sudden increase (see Figure \ref{fig:comparison}). Note that the fitting is performed on the test curve to which we add a point for the chance level (10\% for CIFAR-10) at epoch 0 (see Figure \ref{fig:slowness}). Furthermore, we compute the slowness value only for the fully connected models, as the test curve obtained with the convolutional models is challenging to fit due to a learning rate decay scheme applied already after ten training epochs (see table in Appendix \ref{app:sim_details}
).

The table in Appendix \ref{app:slowness}
shows that the best convergence rate (\textit{i.e.}, small slowness value) is obtained in general by BP, followed by PEPITA. FA has the worst convergence rate. DRTP has a better convergence rate than FA, but it converges to a significantly lower accuracy plateau.

\begin{table*}[h]
\label{tab:results_slowness}
\vskip 0.15in
\begin{center}
\begin{small}
\begin{sc}
\begin{tabular}{l|ccc}
\toprule
& \multicolumn{3}{|c|}{Fully connected models}  \\ \midrule
  & MNIST & CIFAR10 & CIFAR100  \\
\midrule
BP & \textbf{0.026} & \textbf{0.659} & \textbf{1.338} \\ 
 FA & 0.068 & 1.048 & 4.243  \\ 
 DRTP & 0.063 & \textbf{0.336} & 4.05  \\
 PEPITA & 0.049 & 0.785 & 2.552 \\ 
\bottomrule
\end{tabular}
\end{sc}
\end{small}
\end{center}
\caption{Convergence rate in terms of slowness value obtained by BP, FA, DRTP and PEPITA in the experiments for the fully connected models trained on CIFAR-10 (second column of Table \ref{tab:results}). The smallest the slowness value, the better the convergence rate. The slowness is computed on the test curve averaged over 10 independent runs.}
\vskip -0.1in
\end{table*}


\newpage

\section{Error curves for F=0 and for random perturbation}\label{app:F0&RP}
In Figure \ref{fig:F0&RP_main} we show that the update performed by PEPITA is necessary for learning. We compare:
\begin{itemize}
    \item PEPITA, for which, at the modulated pass, the activations are computed as:
        \begin{equation}\label{eq:modulatedpass_orig}
        \begin{split}
            h_1^{err} = \sigma_1 (W_1 (x + F e) ), \\
            h_l^{err} = \sigma_l (W_l h_{l-1}^{err})  \hspace{0.3cm} 2\leq l \leq L.
        \end{split}
        \end{equation}
    \item PEPITA with F=0, for which, at the modulated pass, the activations are computed as:
        \begin{equation}\label{eq:modulatedpass_F0}
        \begin{split}
            h_1^{err} = \sigma_1 (W_1 (x) ), \\
            h_l^{err} = \sigma_l (W_l h_{l-1}^{err})  \hspace{0.3cm} 2\leq l \leq L.
        \end{split}
        \end{equation}
        Therefore, with F=0, only the last layer is trained and the hidden layer is frozen.
    \item Random Perturbation (RP) of the input, for which, at the modulated pass, the activations are computed as:
        \begin{equation}\label{eq:modulatedpass_RP}
        \begin{split}
            h_1^{err} = \sigma_1 (W_1 (x + F \textit{r} k) ), \\
            h_l^{err} = \sigma_l (W_l h_{l-1}^{err})  \hspace{0.3cm} 2\leq l \leq L.
        \end{split}
        \end{equation}
        where $r$ is a random vector with the same dimension as $e$, whose elements are randomly sampled from a uniform distribution over [0, 1), and $k$ is a scalar $\in [0.01,0.1,1]$.
\end{itemize}

$e = h_L - target$ denotes the network error and $F$ denotes the fixed random matrix used to project (\textit{i.e.}, add) the error on the input. \\

Figure \ref{fig:F0&RP_main} shows that PEPITA reaches a significantly higher test accuracy than its variation with F=0 or random perturbation of the input. When F=0, the accuracy saturates at 44.6\%, while PEPITA reaches 52.6\%. For Random Perturbation, if $k=0$ the performance is close to chance level, for smaller values of $k$ the performance is close to that of $F=0$ as the hidden weights are only slightly updated. Hence, the updates obtained by modulating the input with error-related information is key to directing the parameter updates in the correct direction.

\newpage

\section{Analytic results}\label{app:proof}
We observed experimentally that the intrinsic dynamics of PEPITA tend to drive the product between the forward weight matrices to `antialign' with the $F^T$ matrix (Figure \ref{fig:training}b). Here we provide a formal proof showing why such phenomenon occurs in the case of a linear fully connected network.  
We follow the same reasoning presented for feedback alignment in Supplementary Notes 11 and 12 of \cite{Lillicrap} and use the same notation.

We consider a linear network with one hidden layer, where $A$ is the weight matrix from the input to the hidden layer and $W$ is the weight matrix from the hidden to the output layer. Given an input vector $x$, the hidden layer activation is computed as $h=Ax$ (vector), and the output of the network as $y=Wh$ (vector). Each input $x$ is associated with a desired target $y*$ (vector), which is given by a target linear transformation $T$, such that $y* = Tx$. The aim of the training is to learn $A$ and $W$ so that the network is functionally equivalent to $T$. We also define the matrix $E = T - WA$, so that the error vector can be written as $e = Ex$. 

Given a learning rate $\eta$, the weight updates prescribed by PEPITA can be written as:
\begin{align}
    \Delta W = \eta e h^T = \eta E x x^T A^T
\end{align}
and
\begin{align}
    \Delta A &= \eta (h - h^{err}) x^T = \\
    &= \eta (Ax - Ax^{err}) x^T = \\
    &= \eta (Ax - A(x+Fe)) x^T = \\
    &= \eta (Ax - Ax - AFEx)) x^T = \\
    &= -\eta AFExx^T. 
\end{align}
Notice that the prescribed update for $A$ here is computed as $\Delta A = \eta (h - h^{err}) x^T$ rather than $\Delta A = \eta (h - h^{err}) (x+Fe)^T$. As mentioned in Section \ref{sec:learning_rule}, the two rules lead to the same network performance. In this analytical proof for simplicity we consider the first form. \\
We assume that we train the network with batch learning and that the input samples $x$ are i.i.d. standard normal random variables (\textit{i.e.,} mean 0 and standard deviation 1) and thus $xx^T = I$, where $I$ is the identity matrix. Then the parameter updates can be written as:
\begin{align}
    \Delta W = \eta E x x^T A^T = \eta E A^T
\end{align}
and
\begin{align}
    \Delta A = -\eta AFExx^T = - \eta AFE. 
\end{align}
We note that the minus sign in the update of $A$ does not appear in the update of $A$ using feedback alignment \cite{Lillicrap}. Therefore, this step of the proof explains the difference in the direction of the alignment.

In the limit of a small $\eta$, the discrete time learning dynamics converge to the continuous time dynamical system:
\begin{align}
    \dot{W} = EA^T
\end{align}
and
\begin{align}
    \dot{A} = -AFE.
\end{align}
where $\dot{A}$ and $\dot{W}$ are the corresponding temporal derivatives. In order to show why the forward matrices and $F^T$ come to antialignment with each other, we need to prove that the time derivative of $tr(FWA)$ is negative. Indeed, the antialignment is defined by the condition $\frac{d}{dt}tr(FWA)<0$, where $tr(FWA) = [WA]_{\downarrow}\cdot[F^T]_{\downarrow}$ and $[\cdot]_{\downarrow}$ is an operator that flattens a matrix into a vector.
To this goal, we decouple the deterministic dynamics of $A$ and $W$: at first we freeze $W$ and train $A$, and then we keep $A$ constant and train $W$. 

In the first phase, we impose $\dot{W}=0$ and let $\dot{A} = -AFE$. In our experiments we observe that the norm of $A$ monotonically increases during training (Figure \ref{fig:training}c). Thus $\frac{d}{dt}||A||^2 = \frac{d}{dt}tr(A^T A) >0$. We can rewrite the time derivative of the norm as:
\begin{align}
    \frac{d}{dt}tr(A^T A) &= 2 tr(A^T \dot{A}) = \\
    &= 2 tr(A^T (-AFE)) =\\
    &= 2 tr(-A^TAFE) =\\
    &= 2 tr(-||A||^2FE) =\\
    &= -2||A||^2 tr(FE).
\end{align}
Therefore, we deduce $tr(FE)<0$. 

Next, we consider the second phase, in which $\dot{A}=0$ and $\dot{W} = EA^T$. Under this condition, we examine the evolution in time of $tr(FWA)$:
\begin{align}
    \frac{d}{dt}tr(FWA) &= tr(F\dot{W}A) = \\
    &= tr(FEA^T A) = \\
    &= tr(FE||A||^2) = \\
    &= ||A||^2 tr(FE).
\end{align}
From the above result $tr(FE)<0$, we conclude that $\frac{d}{dt}tr(FWA)<0$. This analytical result for a linear network supports our empirical finding on non-linear models that the product of the forward matrices comes to antialign with $F^T$. We conjecture that such antialignment enables the error that modulates the input, and consequently the neural activity, to deliver useful learning signals for weight updates in $A$ and $W$.

\newpage
\section{Weight norm increase}\label{app:wnorm}
Figure \ref{fig:training} shows that the norm of the first-layer weights increase monotonically during the first 50 epochs. Here we report the extended Figure for 1000 training epochs. The bottom panel shows that, after the learning rate decay step at epoch 60, the norm of $W_1$ increases with a significantly smaller rate, however it does not saturate.

\begin{figure*}[h!]
 \centering
 \includegraphics[width=0.5\textwidth]{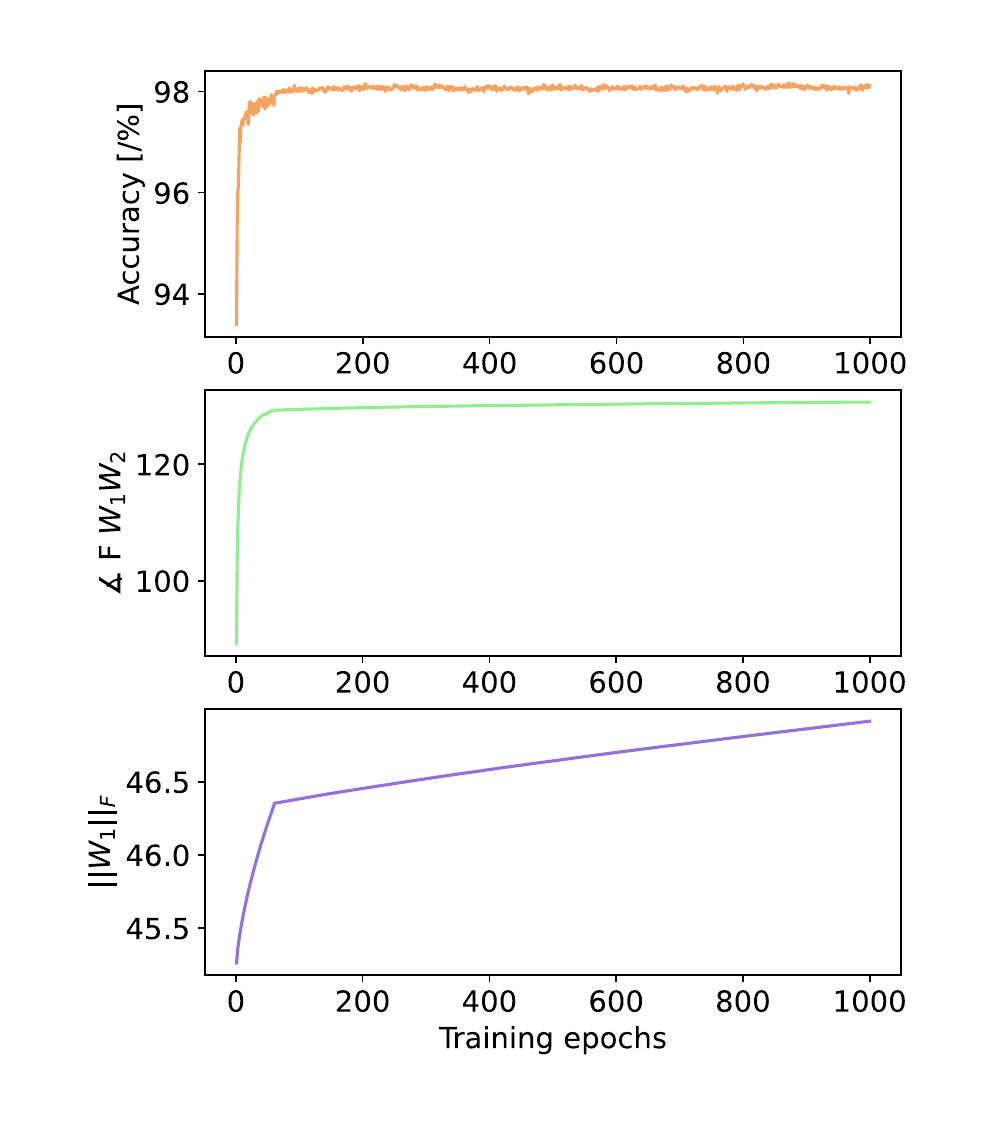}
 \caption{Figure \ref{fig:training} extended to 1000 epochs. Dynamics of a 1 hidden layer network, with 1024 hidden units, 10\% dropout, during training on MNIST with PEPITA. (a) Test accuracy computed after each training epoch. (b) Angle between the product of the forward matrices and the F matrix. (c) Evolution of the norm of the initial weight matrix. The solid line reports the mean over five independent runs, the shaded area indicates the standard deviation.}
 \label{fig:weights1000ep}
\end{figure*}

\newpage
\section{Results for whitened input}\label{app:whiten}
Here we show that introducing the assumption that PEPITA works on whitened input (see Appendix \ref{app:proof}) is plausible. We apply the whitening code proposed in \cite{hadrienj_github_whitening} on MNIST. We use the same settings for the fully connected model as described in Appendix \ref{app:sim_details}, except for the learning rate which needs to be decreased by a factor 10 ($\eta = 0.01$). The Table below reports the accuracy result. The accuracy obtained with whitening is lower than without whitening, however the accuracy is above 90$\%$, showing that the assumption in the anti-alignment proof is plausible.

\begin{table*}[h]
\label{tab:whitened}
\vskip 0.15in
\begin{center}
\begin{small}
\begin{sc}
\begin{tabular}{l|c}
\toprule
  & MNIST  \\
\midrule
 PEPITA non whitened input & 98.01$\pm$0.09\\ 
 PEPITA whitened input & 92.08$\pm$0.14\\ 
\bottomrule
\end{tabular}
\end{sc}
\end{small}
\end{center}
\caption{ Test accuracy [$\%$] achieved by PEPITA with and without whitening on the input with a fully connected model trained on the MNIST dataset. Mean and standard deviation are computed over 10 independent runs.}
\vskip -0.1in
\end{table*}

\newpage
\section{Distribution of trained weights}\label{app:weights}
Here we show that the weight distribution obtained with BP can be fit with a Normal curve, while the distribution obtained with PEPITA is better described by a Student's t-distribution. The Student's t-distribution is a subexponential distribution, which is reminiscent of the connectivity structure of neurons in cortical networks. Indeed, findings of electrophysiological studies indicate that the amplitudes of excitatory post-synaptic potentials between cortical neurons obey a long-tailed pattern, typically lognormal \cite{buszaki,song, Iyer}. 

\begin{figure*}[h!]
 \centering
 \includegraphics[width=0.9\textwidth]{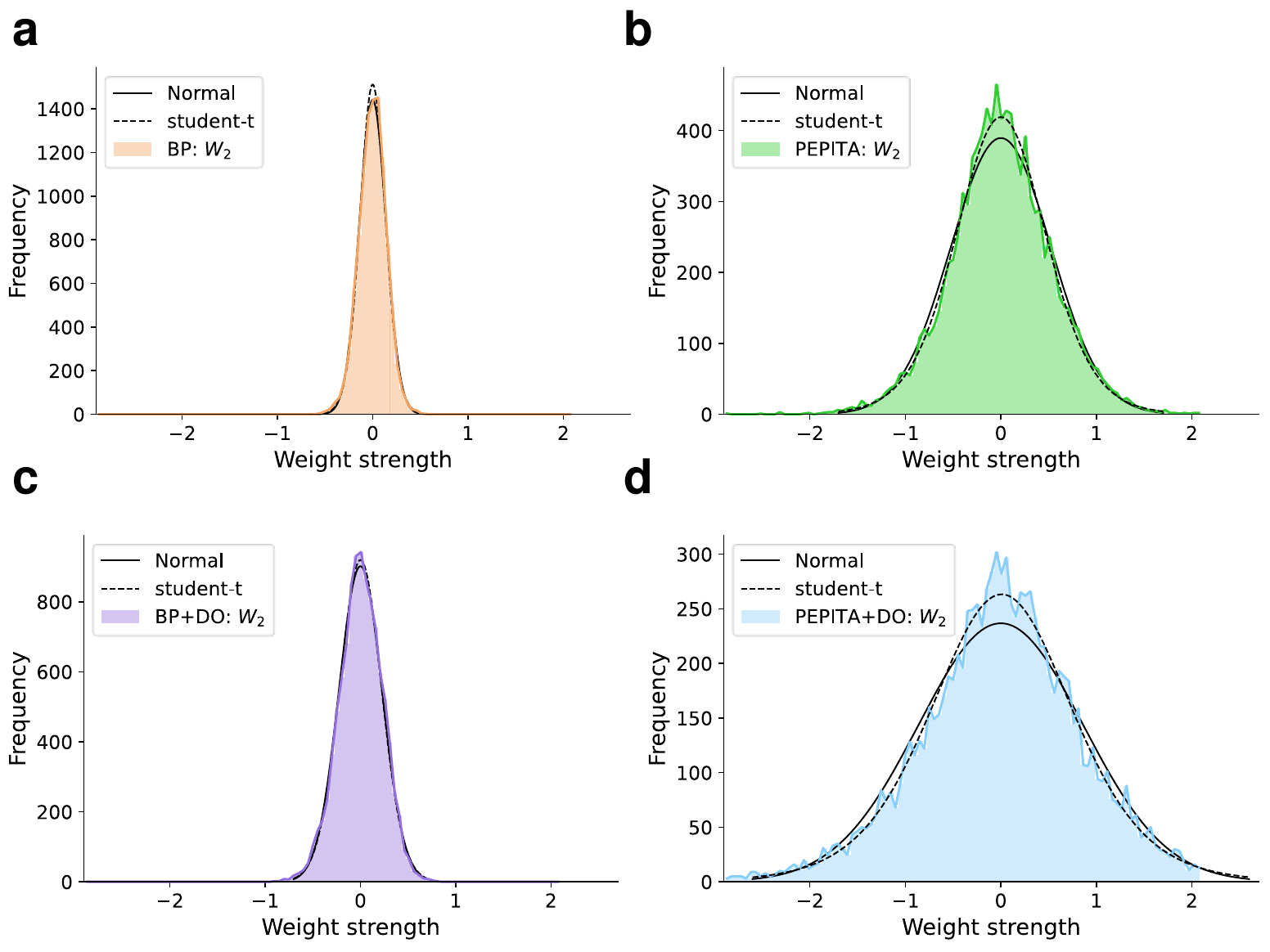}
 \caption{The weight distributions for the output layer reported Figure \ref{fig:weights}b here are shown separately and fit both with a Normal distribution (solid black line) and a Student's t-distribution (dotted black line). With the BP training scheme (a) without dropout and (c) with dropout, the Normal curve well fits the weight distribution. Instead, with the PEPITA training scheme both (b) without dropout and (d) with dropout settings, the Student's t-distribution fits the weight distribution much more accurately than the Normal curve. This indicates that PEPITA pushes the weights towards a sub-exponential distribution. Note that we used different scale for the y-axis for visualization purposes. }
 \label{fig:weights_fit}
\end{figure*}


\end{document}